\documentclass{article}

\usepackage[final]{neurips_2024}
\setcitestyle{numbers,square}

\usepackage{neurips_2024}

\usepackage{graphicx}
\usepackage[utf8]{inputenc} % allow utf-8 input
\usepackage[T1]{fontenc}    % use 8-bit T1 fonts
\usepackage{hyperref}       % hyperlinks
\usepackage{url}            % simple URL typesetting
\usepackage{booktabs}       % professional-quality tables
\usepackage{amsfonts}       % blackboard math symbols
\usepackage{nicefrac}       % compact symbols for 1/2, etc.
\usepackage{microtype}      % microtypography
\usepackage{xcolor}         % colors
\usepackage{xeCJK}
\usepackage{newclude}
 \usepackage{amsmath}
 \usepackage{adjustbox}

\newcommand{\our}{{\tt DynSparse}}

\newcommand{\vQ}{\mathbf{Q}}
\newcommand{\vK}{\mathbf{K}}

\newcommand{\vP}{\mathbf{P}}

\newcommand{\vO}{\mathbf{O}}

\title{
Training-Free Sparse Attention Based on Cumulative Energy Filtering}

\author{
  \small
  Chunlu Li$^{*}$, Yixuan Pan$^{*}$, Bai Du$^{*}$, Zhenyuan Chen$^{*}$, Yanzhao Li$^{*}$, Hui Dong$^{*}$\\
   Hui Wang$^{*}$, Zhiqiang Zou$^{*}$ 
  \\
  \small \texttt * Huawei Technologies\\
}

\begin{document}

\maketitle

\begin{abstract}

    Sparse attention accelerates Diffusion Transformers (DiTs) for video generation by computing only the important tokens while skipping the rest. The token selection strategy is key to balancing sparsity and accuracy. 
    We formulate the token filtering process as a dual-goal optimization problem: maximizing sparsity and minimizing accuracy degradation. Existing algorithms cannot fulfill both objectives simultaneously. For example, Top-p only considers the accuracy constraint, while Top-k maintains a fixed computational budget but loosens the accuracy constraint. 

This paper demonstrates that maintaining a fixed recall rate is sufficient for ensuring accuracy, whereas a fixed threshold is suboptimal for reducing computational cost. Therefore, we propose a dynamic thresholding scheme to improve sparsity while maintaining the same level of accuracy. Furthermore, our algorithm is deeply integrated with Flash Attention (FA), eliminating the need for any additional masking computation overhead.

Experimental results on Wan 2.2 validate that, compared to the BLASST algorithm which is also integrated with FA, our dynamic thresholding strategy enhances sparsity from 61.42\% to 82\% with a VBench metric drop of less than 5\%. This results in an approximate 15\% in attention computation and a $1.61\times$ increase in computational efficiency, which is 1.18x higher than that of BLASST.
  
\end{abstract}

\section{Introduction}
A major limitation of DiT-based vision generation models is that the computational cost of attention scales quadratically with sequence length. Sparse attention has emerged as a mainstream approach for accelerating inference. By filtering out tokens that contribute minimally to the output, it reduces computational overhead while maintaining acceptable accuracy.

  Since Flash Attention 2 \cite{dao2023flashattention} is proposed, attention has been reformulated into tile-based computation by iterating over the combination of $Q_i$ and $K_j$ tiles for better parallelism. Sparse attention leverages this by involving only a subset of these $Q_i$ and $K_j$combinations, skipping the rest to achieve acceleration. Existing methods can be categorized into static and dynamic paradigms, depending on whether the sparsity decisions are determined offline or online.

  \textbf{Sparse attention with static pattern} assumes that each attention head exhibits specific patterns indicating which tokens contribute less to the final attention. 
  Consequently, a set of masks can be predefined to calculate the patterns for each head. During inference, the selected mask or a combination of these masks can be directly applied. SVG \cite{xi2025SVG} first proposes that attention heads can be divided into spatial and temporal heads. The former focuses on the correlations between tokens within a single frame, while the latter pays more attention to the correlation among tokens across temporal frames correspond to the same spatial location. 
  By comparing the errors using different masking patterns, the mask type for each attention head is determined offline. During online inference, the masking pattern is directly applied to determine whether to skip the computation of tokens.
  Methods such as RainFusion and Sparse-VDiT \cite{chen2025rainfusion,chen2026Sparse-vdit} have increased the variety of masks. Some methods have made the process of setting static masks dynamic. For example, Efficient V-DiT \cite{ding2025efficient-vdit} designed parameterized masks, which increased the degree of freedom in mask changes and reduced the loss in accuracy. Other methods have shifted from offline mask selection to online methods. VORTA \cite{sun2025vorta} uses a routing mechanism to select appropriate masks during inference. These static or semi-static methods cannot generalize well on different modes, prompts, layers and diffusion steps.

  \textbf{Dynamic Sparse Attention} To enhance the generalization, dynamic sparse attention make online token selection.
  Dynamic attentions can be further divided into two streams according to the timing of the selection.

  The first category estimates a mask before entering the FA loop. Generally, Q and K are first compressed into Q' and K'. Then, the most important K blocks for each Q block are selected based on a score computed with Q'and K'. The FA inner loop decides whether to skip a tile by querying the mask result. For example, VSA \cite{xi2025SVG} represents Q and K sequences with the mean of Q(K) blocks and use $S' = Q'K'^T$ to approximates the S matrix. Then, it keeps the Top-K largest elements and skip the rest blocks.

  Feature compression reduces the overhead of mask computation. The most widely used is the block mean. Other variants include: X attention \cite{xu2025xattention} uses the average value of the diagonal elements in a block to represent the entire block. The DSV algorithm \cite{tan2025dsv} projects a sequence into a low-dimensional space using a learnable low-rank matrix.
  To make the compressed features more representative, some methods use reordering to group similar tokens into the same block. Currently, the mainstream is to rearrange the 1D sequence to the time-space dimension of [h, w, t], e.g. PAROAttention, BSA, etc. \cite{zhao2025paroattention,team2025longcat,zhan2025bidirectional}. This method groups tokens that are close in time and space together so as to reducing the variance within a block.
  SVG2 \cite{yang2025SVG2} proposes to use K-means for clustering and use the center to represent each cluster. Sparge \cite{zhang2025spargeattention} employs Hilbert reordering and uses the cosine similarity to determine whether the average value is representative. For blocks that cannot be represented by the average value, all tokens are selected for further attention computation.

  The most prevalent element selection methods are Top-K and Top-P. The Top-P algorithm performs computation on an approximated P' matrix. This algorithm first sorts the elements in descending order and then truncates the distribution at the point where the cumulative probability surpasses a specific threshold..
  Methods such as SVG2 and XAttention use Top-P on the $\vP$ matrix, while methods such as SLA, BSA in LongCat, and AdaSpa use Top-K on the $\vQ\vK^T$ matrix. In addition to the Top-P and Top-K algorithms, the BSA algorithm \cite{zhan2025bidirectional} calculates a dynamic K based on statistical values to improve the accuracy. FlashPrefill \cite{fan2026flashprefill} uses the weighted maximum $\alpha max(\vQ\vK^T)$ as the threshold to filter out unimportant elements. SSTA \cite{wu2025hunyuanvideo} in HunyuanVideo 1.5 is based on the prior importance of the elements on the main diagonal of the QKT matrix. It integrates the SWA with Top-K to prevent missing important elements.

  The second category of dynamic sparsity algorithms decides whether to skip certain blocks within the FA loop. A typical example is the BLASST algorithm \cite{yuan2025blasst}. In the inner loop, $\hat{m}_i - m_{i}^{(j)}$ is used as an indicator of the importance of each tile, where $\hat{m}_i$ is the maximum of the current tile and $m_{i}^{(j)}$ is the running max. When this indicator is below a certain threshold, subsequent computations are skipped. The essence of this method is similar to thresholding with the weighted maximum $\alpha max(\vQ\vK^T)$, while the global max is replaced by running max for engineering convenience. This approach does not require additional mask computation and can reduce the overall computational complexity.
  LiteAttention further incorporate a mask reuse mechanism to further reduce the cost of threshold judgment. Compared to the first category of dynamic sparse attention, it requires no mask computation overhead.

  The core of dynamic sparsity is how to select appropriate tokens to ensure low computational cost without compromising accuracy.
  However, existing algorithms cannot fulfill the two objectives simultaneously. The widely adopted Top-K strategy maintains a fixed computational budget but loosens the accuracy constraint. Conversely, Top-P only considers the accuracy and ignores the former goal.

    This paper demonstrates that maintaining a fixed recall rate is sufficient for ensuring accuracy. However, it is not optimal for reducing computational cost. Starting with the dual-goal optimization problem, we propose a dynamic threshold calculation scheme based on cumulative energy, i.e. \our, that improves sparsity under the same accuracy. Besides, our algorithm is deeply integrated with the FA algorithm and does not introduce extra computation overhead.

  Specifically, we propose thresholding the tokens based on the accumulated energy $\ell_{i}^{(j)}$. In FA, $\ell_{i}^{(j)}$ represents the probability of selected tokens accumulated so far. It implies the upper bound of the potential contribution of subsequent elements to the softmax-normalized $\vP$ matrix. Consequently, it serves as a more effective metric for assessing the importance of incoming tokens. Furthermore, the cumulative nature of $\ell_{i}^{(j)}$ ensures that as its value increases, fewer tokens are selected, thereby enhancing sparsity. Similar to BLASST, this method is deeply integrated with FA. It is also worthy to note that $\ell_{i}^{(j)}$ is contained naturally in the FA operator, thereby minimizing additional overhead.

  Experiments on Wan 2.2 verifies that, our training-free method can achieve a sparsity of 82.14\% on a 14B model with block size 128, representing an improvement of 20.72\% compared to the BLASST baseline. On a 5B model, our method can achieve a sparsity of 81.38\%. The ablation study also validates the robustness of the proposed method.

\section{Sparse Attention Anaysis}
\label{Section:Analysis}

\subsection{Sparse Attention as an Optimization Problem}

We first formulate the sparse attention as an optimization problem and explain the defect of existing methods. Through derivation, we then verified that the selection criterion, $\Delta P < \epsilon$, is a sufficient condition to ensure accuracy. However, to ensure optimal sparsity $\epsilon$ varies under different P matrix distributions.

\subsubsection{Problem Statement}
\label{Sec2: optimal epsilon}

The objective of sparse attention can be defined as: $$min(cost_{compute}), s.t. dist(\vO,\vO') <\theta$$, where $\vO$ and  $\vO'$ are the outputs of full attention and sparse attention. $dist$ denotes the error between the two inputs.

Assume the elements $v_j$ of $\mathbf{V}$ are i.i.d and employ $|\vO-\vO'|$ to reflect the error.
Let $S_1 = \sum_{j \in \mathcal{S}} p_j \le \epsilon$, where $\mathcal{S}$ denotes the set of discarded elements. The error variable $E$ consists of two parts:
\begin{equation}
E = \underbrace{\sum_{j \in \mathcal{S}} p_j v_j}_{\text{loss due to discarding}} - \underbrace{\frac{S_1}{1 - S_1} \sum_{j \in \mathcal{K}} p_j v_j}_{\text{amplified bias due to renormalization}}
\label{eq:sparse attn error}
\end{equation}
The existing methods are all simplifications of the optimization objectives.

1) Top-P: Assume that $v_j$ has an upper bound $|V|$.
By ensuring $\sum_{j \in \mathcal{S}} p_j < \epsilon$, it can be guaranteed that $|\vO - \vO'| < \epsilon |V|$. However, this method does not ensure least computational cost.

2) Top-K maintains a fixed computational budget and loosens the accuracy constraint. 

3) Truncating tokens below the weighted P matrix maximum \cite{fan2026flashprefill} cannot guarantee optimal accuracy and sparsity using a fixed threshold weight.
Assume the sequence length is $L$,$S=\{P_j|P_j<\alpha P_{max}\}$ are the skipped K tokens. the sum of these retained tokens are $\gamma P_{max}$, retained probability sum is $\sum_{j \notin C}=\gamma (N-K)P_{max}$.
To determine the upper bound of the precision error, we assumed that the discarded tokens share the same value of $\alpha P_{max}$. In such situation, to achieve the maximum sparsity while satisfying the precision error, it is required that: $\alpha = \frac{\sum_{i \in S}{p_i}}{1-\sum_{i \in S}{p_i}} \frac{L-K}{K}$. It suggests that the optimal threshold $\alpha$ is related to the distribution of the $\vP$ matrix. Since the $\alpha$ used in this method is generally fixed, the sparsity and precision may not meet expectations in certain cases.

4) \textbf{BLASST} replaces the global max with a running max for engineering optimization so as to integrate deeply with FA. Since the running max may include some tokens with lower importance that do not contribute substantially to accuracy when it has not yet reached the global maximum, Relative Thresholding serves as the upper bound for the accuracy-sparsity tradeoff in BLASST. The theoretical accuracy-sparsity upper bound is the same as using the global max.

\subsection{Theoretical Derivation of Optimization Objectives}
First, we verify that the Top-P strategy, which uses the recall rate as a threshold，i.e. $\Delta P<\epsilon$, for selection can strictly ensure the upper bound of error yet $\epsilon$ should vary according to the distribution of the $\vP$ matrix. Note that ensuring a lower sparsity equivalents to selecting a higher $\epsilon$.

The mathematical expectation of the error in Eq. \ref{eq:sparse attn error}, \(\mathbb{E}[E] = 0\), means that performing Softmax re-normalization is an unbiased truncation operation.
The\textbf{ variation} can be formulated as
\begin{equation}
\text{Var}(E) = \sigma^2 \left[ \sum_{j \in \mathcal{S}} p_j^2 + \left( \frac{S_1}{1 - S_1} \right)^2 \sum_{j \in \mathcal{K}} p_j^2 \right]
\end{equation}

According to the Central Limit Theorem or Chebyshev's inequality, the precision requirement can be rewritten as: $\mathbb{E}[E]+\text{Var}(E)=\text{Var}(E) \le \delta \cdot \theta^2$, where $\delta$ is a constant related to the confidence level. Substituting $S_1 = \epsilon$ into the variance formula, we obtain the constraint:

\begin{equation}
\sum_{j \in \mathcal{S}} p_j^2 + \frac{\epsilon^2}{(1 - \epsilon)^2} \sum_{j \in \mathcal{K}} p_j^2 \le \frac{\delta \cdot \theta^2}{\sigma^2} =\Theta 
\label{eq:theoratical optimal epsilon}
\end{equation}

Since $\epsilon$ is small, a first-order Taylor approximation gives $\frac{\epsilon}{1 - \epsilon} \approx \epsilon$. Moreover,
we denote the squared 2-norm of the vectors as $\|\mathbf{P}_{\mathcal{S}}\|_2^2 = \sum_{j \in \mathcal{S}} p_j^2$ and $\|\mathbf{P}_{\mathcal{K}}\|_2^2 = \sum_{j \in \mathcal{K}} p_j^2$,
and $\epsilon_{ub}$ is the theoretical upper bound of $\epsilon$, satisfying the following condition:
\begin{equation}
\|\mathbf{P}_{\mathcal{S}}\|_2^2 + \epsilon_{ub}^2 \|\mathbf{P}_{\mathcal{K}}\|_2^2 = \Theta
\label{Eq: theoretical upper bound}
\end{equation}

Eq. \ref{Eq: theoretical upper bound} indicates that the optimal value of $\epsilon$ is related to the distribution of $\mathbf{P}$ . $\sum p_j^2$ (i.e., the squared L2 norm) statistically measures the sharpness of the distribution. We discuss two extreme cases:

Extremely flat distribution:
The probability of discarding tokens reaches $\epsilon$.
However, $\|\mathbf{P}_{\mathcal{S}}\|_2^2$ and $\|\mathbf{P}_{\mathcal{K}}\|_2^2$ will be extremely small. For example, under a uniform distribution, the upper bound of both is $1/L$, where $L$ is the sequence length, which is typically on the order of 1e5 in video generation.
In this case, even if you choose a larger $\epsilon$, the error $|E|$ will not exceed the threshold.
Therefore, you can set a larger $\epsilon$ and aggressively discard more tokens.

For peaky and concentrated distribution, the attention is highly concentrated on a few tokens, and $\|\mathbf{P}_{\mathcal{K}}\|_2^2$ is large.
In this case, even a slightly larger $\epsilon$ can cause the output of the retained tokens after renormalization to fluctuate beyond the threshold $\theta$. For example, when 1k tokens share almost 99\% of the attention, $\|\mathbf{P}_{\mathcal{K}}\|_2^2 \approx 1e-3$.
Therefore, under a sharp distribution, the penalty for renormalization is significant, and a very small $\epsilon$ (or even no discarding) must be set.
In summary, $\epsilon$ is highly correlated with the distribution of the $\mathbf{P}$ matrix.

Using a fixed $\epsilon$ can strictly ensure the accuracy error, but it cannot achieve the optimal sparsity. Strictly ensuring $\Delta P < \epsilon$ is a sufficient condition for accuracy, but not a necessary one.
In Sec. \ref{Appd: Accuracy vs Recall}, we used recall values on Wan 2.2 to reflect the value of $\epsilon$ and compared the relationship between MAE and recall under different distributions, validating the above analysis.

\subsection{Sources of Computation Reduction}
According to Sec. \ref{Sec2: optimal epsilon}, when the distribution is relatively flat, a more aggressive value of $\epsilon$ can be set. To analyze the theoretical benefits in this case, we assume
\textbf{P tends toward a uniform distribution.}
Assuming $K$ elements are retained, and $\delta$ is the standard deviation of V, the upper bound of the error in the standard deviation is:
$$std = \frac{\delta}{\sqrt K} \sqrt{\frac{N-K}{N-1}} $$
For Wan 2.2 5B, $N_{seq}=27280$, with a block size of 128. When $N_{blk} = ceil(N_{seq}/128)=214$, only $80\%$ of the blocks need to be retained, and the std error is 3.4\% of $\delta$. When 95\% is retained, the std error is 1.6\% of $\delta$. It can be seen that when sparsity increases from 0.05 to 0.2, the std difference is approximately 2 times, with no significant difference in the error distribution.
As the sequence length N increases, this sparsity improvement becomes more pronounced. For example, in Wan 2.2 14B, $N_{seq}=75600$, $Bc=128$, and the standard deviation difference is acceptable after sparsity increases from 0.05 to 0.6.

\section{Method}

Section \ref{Section:Analysis} indicates that filtering tokens (blocks) using $\sum p_j < \epsilon$ could not reach optimal sparsity with a fixed $\epsilon$. Instead, the selection should follow Eq. \ref{Eq: theoretical upper bound} in which $\epsilon$ should be set to its upper bound $\epsilon_{ub}$. The process of mask computation should follow:

Given an upper tolerance limit for the error variance $\Theta \propto \theta^2 / \sigma^2$.
Sort $P_j$ in ascending order.
Initialize $\mathcal{S} = \emptyset$, $\mathcal{K} = \{1 \dots N\}$.
Sequentially move the smallest $p_j$ from $\mathcal{K}$ to $\mathcal{S}$, and dynamically update $S_1 = \sum_{\mathcal{S}} p_j$. Calculate the current error variance metric in real time, and sTop-the computation when the condition represented by Eq. \ref{eq:theoratical optimal epsilon} is met.

Although this method can achieve the desired accuracy and optimal sparsity, it requires softmax computation, sorting, and dynamic decision-making before FA. Therefore, it is difficult to achieve significant performance gains in practice.

\subsection{\our{}}

We propose a metric for selecting tokens, similar to the BLASST algorithm, which can filter tokens during the FA process:
\begin{equation}
\label{eq:thresholding matrix}
t_i^j=\hat{m}_i - \mathrm{ln}{\ell_{i}^{(j-1)} }- m_i^{(j-1)} = \hat{m}_i - \text{LSE}_i^{(j-1)}
\end{equation}
Here, $\hat{m}_i$ represents the local maximum value of each tile, and $m_i^{(j-1)}$ represents the cumulative maximum value. The cumulative energy (Log-Sum-Exp) after $(j-1)$ blocks is:
\begin{equation}
    \text{LSE}_i^{(j-1)} = \ln \left( \sum_{k \leq j-1, k \notin \mathcal{J}_i}  e^{S_{i}^k - m_i^{(j-1)}} \right) + m_i^{(j-1)} = \ln \ell_i^{(j-1)} + m_i^{(j-1)}.
\end{equation}
We use $t_i^j$ as the decision metric and define a global threshold $\lambda$. If $t_{i,j} < \lambda$, the $j$-th block is considered energetically insignificant and thus its computation is skipped.

First, we verify that this method can strictly guarantee the constraint $\Delta recall = \Delta P < \epsilon$. For a single tile $i$ that is about to be skipped, its contribution to $\Delta recall$ is as follows:
$$\Delta recall_i = \frac{e^{A_i}}{ \sum_{i \in [1,L]}e^{A_i}} \leq \frac{e^{\hat{m}_i}}{\sum_{i \in \mathcal{K}}e^{A_i}} = e^{t_i}$$
Since our screening condition constrains $t_{i} < \lambda$, the contribution of each token to the recall is $recall \leq e^{\lambda} < \epsilon$.
In the worst-case scenario, assuming that the error contribution of each token reaches the upper bound of the error, we can set $\lambda = \ln(\epsilon/L)$ to ensure that the constraint is strictly not violated. This worst-case scenario corresponds to the case where the P matrix is uniformly distributed.
Combining the analysis from the previous section, we can ensure that this method guarantees the sufficient condition for accuracy, $\Delta P < \epsilon$, based on this.

Further, we explain how \our{} ensures optimal sparsity under the condition of maintaining accuracy. In high-entropy distributions (uniform distributions), as the sequence length increases, LSE (log-sum estimate) also increases, making the threshold $\hat{m}_i^{(j)} < \text{LSE}_i^{(j-1)} + \lambda$ increasingly stringent, thereby allowing aggressive pruning of dilution blocks that do not significantly alter the attention output. In low-entropy (spiky) distributions, once a global maximum ("winner") is encountered, LSE rises sharply, causing all subsequent low-scoring blocks' $t_{i,j}$ to immediately fall below $\lambda$, triggering early exit.

It is worth noting that our method does not require additional computation to estimate the mask, as all variables ($\hat{m}_i^{(j)}, \ell_i^{(j-1)}, m_i^{(j-1)}$) are already maintained by the online softmax algorithm in FA. % 

\textbf{Hyperparameter Selection}
The threshold $\lambda$ has a clear physical interpretation: it represents the logarithm of the ratio of the potential energy of a tile to the accumulated mass.
A reasonable starting point is to assume a uniform distribution under maximum entropy, where the expected relative contribution of each tile is $B_c/L$. We suggest setting $\lambda$ slightly above this "noise floor" value to distinguish informative tiles from negligible ones:
In practice, we recommend:
\begin{equation}
\lambda = \ln(P_{real}) + \ln \left( \frac{B_c}{L} \right),
\end{equation}
where $P_{real}$ is a calibration constant that acts as a significance multiplier (peak-to-average ratio).
This formula takes into account the "heavy-tailed" nature of attention distributions in large-scale models. Specifically, $ln(P_{real})$ determines the position of the pruning boundary within the distribution:
\textbf{High-amplitude regions ($P_{\text{real}} \gg 2$):} Retain dominant "heavy-head tiles" that define semantic alignment.
\textbf{Transition regions ($P_{\text{real}} \approx 1.5$):} Capture moderately informative tokens that preserve fine-grained structural details.
\textbf{Noise tails ($P_{\text{real}} < 1$):} Prune tokens with energy below the average expectation, which are likely to be unimportant.
For the Wan 2.2 14B model ($L=75600, B_c=128$), setting $\lambda$ to -6 provides a robust balance, significantly reducing FLOPs across varying content densities while maintaining high-quality generation.
Furthermore, mapping the one-dimensional sequence back to its original spatiotemporal dimensions before processing (as done in other sparse attention methods \cite{zhao2025paroattention,wu2025hunyuanvideo}) can enhance the correlation within tiles.

\section{Experimental Results}

\subsection{Settings}

\smallskip
\noindent{\textbf{Baseline.} 
For video generation models, we consider Wan2.2-5B and Wan2.2-A14B~\citep{wan2025wan}. 
FlashAttention3 (FA3)~\citep{shah2024flashattention}, SparseVideoGen (SVG)~\cite{xi2025SVG}, RadialAttention (Radial)~\cite{li2025radial}, LiteAttention, BLASST comparison baseline. 

% \smallskip

Generated video quality is evaluated using VBench~\cite{huang2024vbench} across the metrics \textit{Aesthetic Quality} (AQ), \textit{Background Consistency} (BC), \textit{Dynamic Degree} (DD), \textit{Imaging Quality} (IQ), \textit{Subject Consistency} (SC), \textit{Temporal Flickering} (TF), and \textit{Temporal Style} (TS). 
All values are averaged over the dataset.
For \our{}, \textit{sparsity} (Sps) denotes the fraction of computations skipped relative to full attention, averaged over the generation process. 
For all other methods, we report the sparsity values they report.

\subsection{Quantitative Results}

\subsubsection{Effectiveness and Efficiency}

\begin{table*}[h!]
\small
\setlength{\tabcolsep}{3pt} %

\centering
\caption{Comparison of \our{}'s video quality (VBench), attention sparsity (Sps), and runtime (Run) compared with FlashAttention3 (FA3), SparseVideoGen, and RadialAttention over Wan2.2-5B and Wan2.2-14B.
Best results are in \textbf{bold} and second best in \textit{italic}.}
\begin{tabular}{lccccccccccc}
\toprule
\textbf{Wan2.2-14B} \\
\midrule
 \textbf{Method} &\textbf{AQ} $\uparrow$   & \textbf{BC}  $\uparrow$  & \textbf{DD}  $\uparrow$  & \textbf{IQ}  $\uparrow$  & \textbf{OC} $\uparrow$   & \textbf{SC}  $\uparrow$  & \textbf{TF}  $\uparrow$  & \textbf{TS}  $\uparrow$  & \textbf{Sps}$\uparrow$ & \textbf{SpeedUp}$\uparrow$\\

\midrule

FA3           & 0.695 & 0.976 & 0.807 & 0.709 & 0.278 & 0.911 & 0.986 & 0.255 & 0  &   1 \\
SVG             & {0.689} & 0.962 & {-} & {-} &-& - & {0.952} & - & {66} &1.44\\
Radial          & 0.682   & {0.974} & {-} & {-} &-& {-} & 0.947 & - & {66} &1.22\\

Lite            & \textbf{{0.698}} & {0.977} & {-} & - &-& {-} & {0.953} & - & {32}& \textbf{1.65} \\
BLASST             & 0.657 & 0.000 & 0.571 & \textbf{0.693} & 0.251 & \textbf{0.888} & \textbf{0.988} & \textbf{0.262} & 61.42 & 1.32 \\

\our{}             & \textit{0.697} & 0.961 & \textbf{0.786} & 0.665 & \textbf{0.276} & 0.879 & 0.981 & 0.253 & \textbf{82.14} & \textbf{1.61} \\

\toprule
\textbf{Wan2.2-5B} \\
\midrule
 \textbf{Method} &\textbf{AQ} $\uparrow$   & \textbf{BC}  $\uparrow$  & \textbf{DD}  $\uparrow$  & \textbf{IQ}  $\uparrow$  & \textbf{OC} $\uparrow$   & \textbf{SC}  $\uparrow$  & \textbf{TF}  $\uparrow$  & \textbf{TS}  $\uparrow$  & \textbf{Sps}$\uparrow$ & \textbf{SpeedUp}$\uparrow$\\

\midrule

FA3             & 0.666 & 0.966 & 0.638 & 0.700 & 0.279 & 0.936 & 0.994 & 0.258 & 0  &1   \\
\midrule
Top-K ($K=0.8$) & \textbf{0.656} & 0.961 & \textcolor{red}{0.571} & 0.674 & 0.279 & 0.896 & 0.992 & 0.256 & 0.8006 & - \\
Top-P ($P=0.9$)         & \textit{0.655} & 0.950 & \textbf{0.750} & 0.670 & 0.278 & 0.905 & 0.990 & 0.257 & 0.7958 & - \\
BLASST                & 0.651 & \textit{0.967} & 0.667 & \textbf{0.695} & 0.255 & \textbf{0.905} & \textbf{0.995} & \textbf{0.250} & 74.41 &  \textbf{1.41}\\
\our{}                & 0.652 & \textbf{0.972} & \textit{0.695} & 0.662 & \textbf{0.283} & \textbf{0.905} & \textit{0.993} & 0.248 & \textbf{81.38}  & \textbf{1.61}  \\

\bottomrule
\end{tabular}
\label{Tab:VBench-quality assessment}
\end{table*}

\smallskip
The metrics on VBench are presented in Tab.\ref{Tab:VBench-quality assessment}. The numerical results for SVG Liter and Radial are excerpted from the Lite paper. Since the values of DD, IQ, OC, SC, and TS in the FA3 baseline vary significantly, their data are not used.

The table shows that the method proposed in this paper can ensure that the quality metric loss remains within an acceptable range of less than 5\% for sparsity levels exceeding 80\%. For the Wan 2.2 14B model, when the sparsity level of the BLASST algorithm exceeds 60\%, the AQ, DD, and OC metrics drop significantly, exceeding the acceptable range.
Therefore, it can be concluded that compared to the BLASST algorithm, our dynamic thresholding approach can increase the sparsity level from 61.42\% to 82.14\% on the 14B model.
For the 5B model, we increased the sparsity level from 74.41\% to 81.38\%. This is because the sequence length is shorter and the information redundancy is lower, thus the sparsity improvement is more limited.
We set a block size of 128 for the 14B model and 32 for the 5B model to reach comparable sparsity.

At similar sparsity levels, Top-P ($p=0.9$) and Top-K ($k=0.8$), although the metrics of our method are lower, the DD of Top-P is abnormally high. This is because the generated frames by these two algorithms exhibit jitter, resulting in poor video quality. For more details, refer to Sec. \ref{Sec:qualitative comparison} and the results in Fig. \ref{figure:visualize - frames} in the Appendix.
The abnormally low DD of Top-K also reflects the insufficient quality of the generated videos.

\subsubsection{Sparsity across timesteps and layers}
\begin{figure}[h!]
  \centering
  \includegraphics[width=\textwidth]{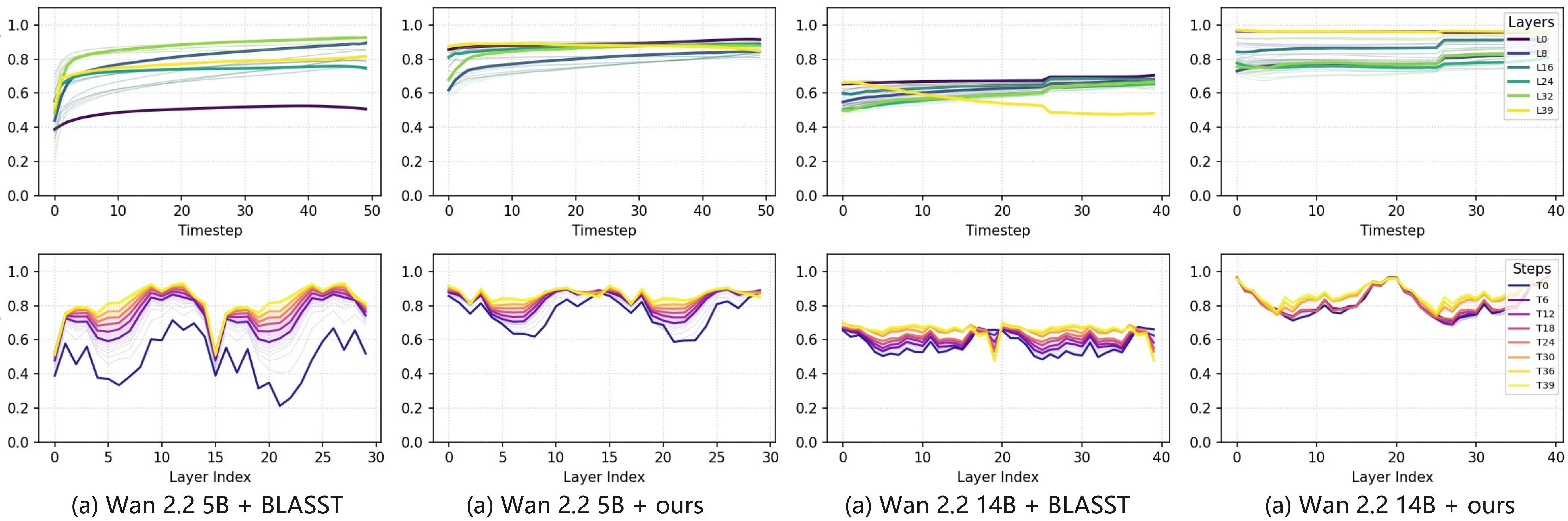}
  \caption{Sparsity across timesteps and layers.}
  \label{figure:sparsity}
  
\end{figure}

Fig. \ref{figure:sparsity} shows the sparsity of BLASST and our method under the same prompt, as a function of timestep and layer, for the Wan 2.2 5B and 14B models. Overall, the sparsity of \our{} is higher than that of BLASST, and the variance in sparsity between different layers (at different timesteps) is reduced.
It is worth noting that, whether using the BLASST or \our{} algorithm, certain layers of Wan2.2 exhibit lower sparsity. For example, layers 5-8 and 20-25 in the 5B model, and layers 5-10 and 25-30 in the 14B model. This phenomenon may indicate that these layers are more sensitive to accuracy.
Similarly, the denoising process is more sensitive to accuracy in the early stages and becomes more robust in the later stages.

\subsection{Qualitative Results}
\label{Sec:qualitative comparison}

\begin{figure}[h!]
  \centering
  \begin{adjustbox}{trim={0.1\width} 0 0 0, clip}
  \includegraphics[width=1.1\textwidth]{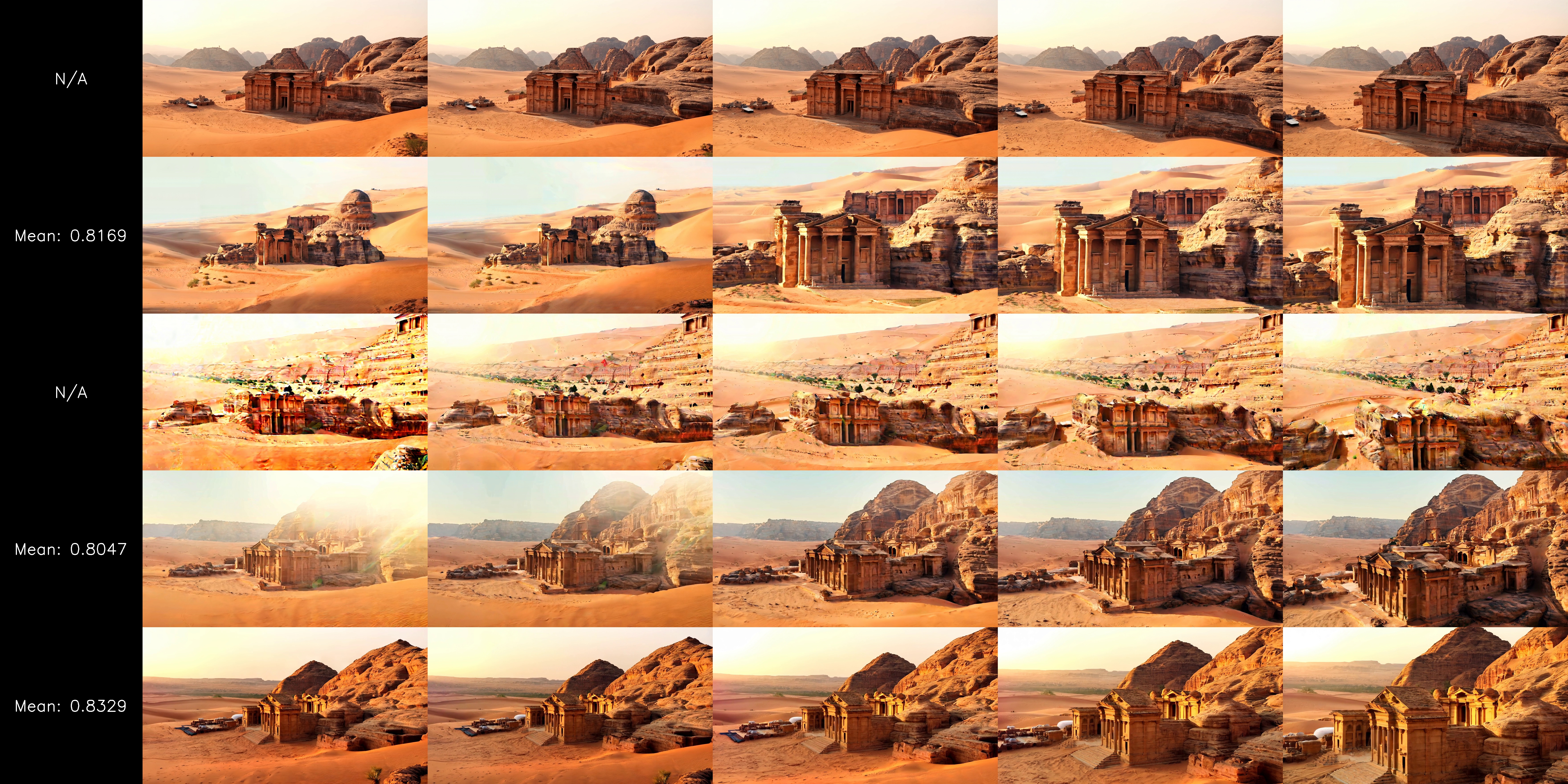}
  \end{adjustbox}
  % \fbox{\rule[-.5cm]{0cm}{4cm} \rule[-.5cm]{4cm}{0cm}}
  \caption{Prompt:A tranquil tableau of at the edge of the Arabian Desert, the ancient city of Petra beckoned with its enigmatic rock-carved facades. From Top-to bottom: FA3, Top-P (sparsity 0.7732), Top-K (0.8), BLASST (sparsity 78.5\%), and \our{} (sparsity 82.1\%).}
  \label{figure:visualize - video 1}
\end{figure}

Fig. \ref{figure:visualize - video 1} shows the screenshots of the generated videos, with the results of FA3, Top-P, Top-K, BLASST, and \our{} displayed from Top-to bottom. As shown in Fig. \ref{figure:visualize - video 1}, at similar sparsity levels, Top-P, Top-K, and BLASST exhibit some artifacts. Additionally, the camera movement in the video generated by Top-P is rapid. In contrast, the artifacts in the \our{} results are fewer, and the camera movement is more similar to that of the FA3 baseline.
Fig. \ref{figure:visualize - frames} in the appendix reveals more visual results.

\subsubsection{Visualization of the Masks}
\begin{figure}[h!]
  \centering 
  \small
  \includegraphics[width=\textwidth]{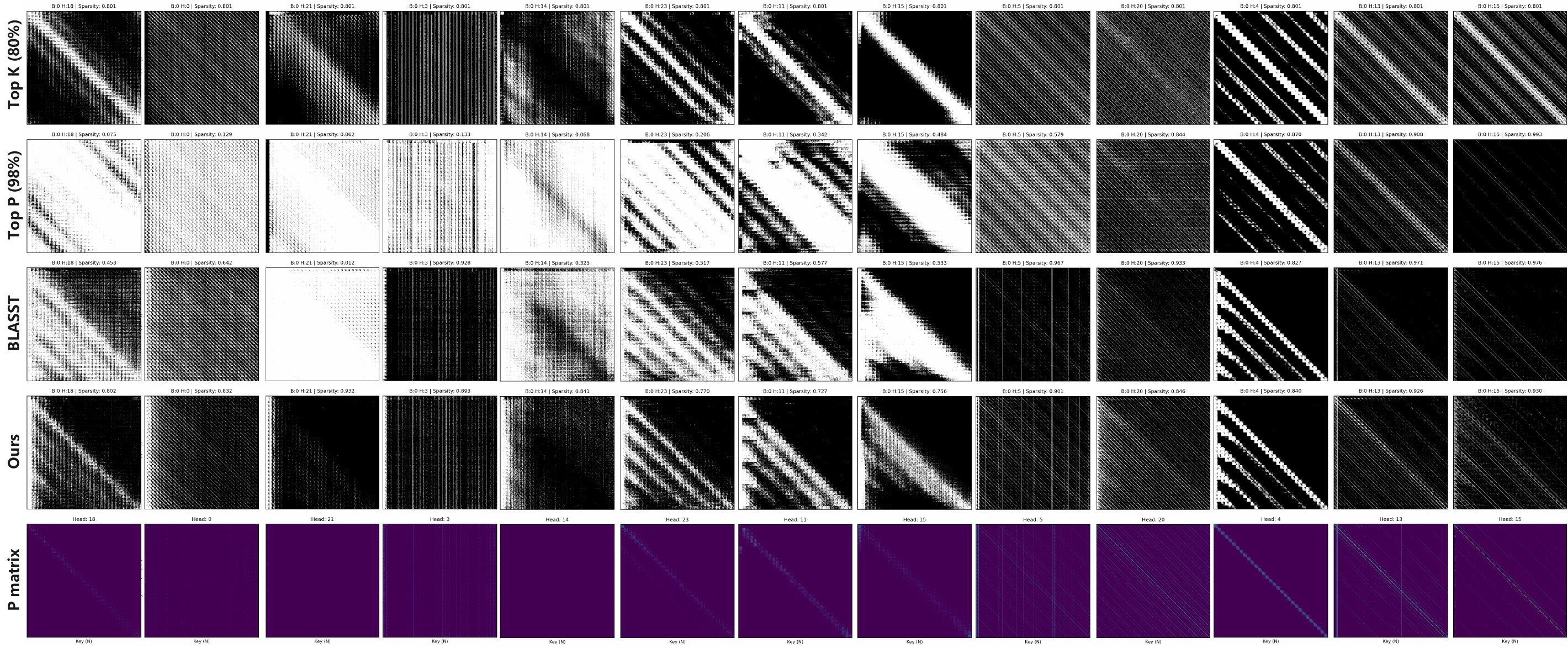}
  \caption{
  Comparison of mask differences under different P matrix distributions using various methods. From Top-to bottom, the results are the mask calculations for Top-K, Top-P, BLASST, and the method proposed in this paper. The last row shows the visualization results of the P matrix.}
  \label{figure:masks}
  
\end{figure}

We visualize the masks of different methods to compare the differences between them.
Fig. \ref{figure:masks} shows the first five columns where all activation values are similar. In this case, the Top-P method needs to retain more elements to ensure that the cumulative cdf value exceeds the threshold. For the BLASST method, its mask depends not only on the distribution of the P matrix elements but also on the position of the maximum value. For example, in the 4th column, the maximum value in each row of the P matrix appears at the first token position, so the subsequent selection ratio drops significantly, while the 3rd column shows the opposite. In contrast, our method can effectively select the most important elements.

The 6th to 9th columns show cases where some elements have larger activation values. In this case, the sparsity of the Top-P method improves. For the BLASST algorithm, since the maximum value in each row is on the diagonal, BLASST retains more elements in the lower-left corner. In contrast, \our{} achieves faster LSE accumulation during the computation, thus retaining fewer elements in the lower-left corner and the right side.

The last four columns show cases with fewer extreme values in the activation values. In this case, most methods perform effectively. The redundancy of Top-K is significantly reduced. BLASST and \our{} show similar performance.

Overall, under the same precision, \our{} achieves better overall sparsity in the mask. The experimental results are consistent with the analysis in Sec. \ref{Section:Analysis}.

\subsection{Ablation Study}
\label{Sec:Ablation}

\begin{table*}[h!]
\small
\setlength{\tabcolsep}{3.5pt} % 
\centering
\caption{Comparison of \our{}'s video quality (VBench), attention sparsity (Sps), and runtime (Run) compared with FlashAttention3 (FA3) over Wan2.2-5B.
Best results are in \textbf{bold} and second best in \textit{italic}.}
\begin{tabular}{lccccccccccc}
\toprule
\textbf{Method} & \textbf{AQ}$\uparrow$ & \textbf{BC}$\uparrow$ & \textbf{DD}$\uparrow$ & \textbf{IQ}$\uparrow$ & \textbf{OC}$\uparrow$ & \textbf{SC}$\uparrow$ & \textbf{TF}$\uparrow$ & \textbf{TS}$\uparrow$ & \textbf{Sps[\%]}$\uparrow$ \\
\toprule
\textbf{Wan2.2-5B} \\
FA3             & 0.666 & 0.966 & 0.638 & 0.700 & 0.279 & 0.936 & 0.994 & 0.258 & NaN      \\
\midrule
-6                & 0.355 & 0.961 & 0.714 & 0.521 & 0.161 & 0.840 & 0.963 & 0.151 & 0.9211   \\
-8                & 0.620 & 0.943 & 0.500 & 0.630 & 0.232 & 0.864 & 0.994 & 0.209 & 0.8533   \\
-9                & 0.652 & 0.972 & 0.695 & 0.662 & 0.283 & 0.905 & 0.993 & 0.248 & 0.8138   \\
-11               & 0.685 & 0.967 & 0.694 & 0.661 & 0.273 & 0.927 & 0.994 & 0.255 & 0.6786   \\
-13               & 0.647 & 0.973 & 0.600 & 0.682 & 0.254 & 0.917 & 0.997 & 0.259 & 0.5076   \\
\bottomrule
\end{tabular}
\label{Tab:VBench ablation}
\end{table*}

Table \ref{Tab:VBench ablation} shows the results of our method with different parameter settings. It can be seen that the sparsity increases as the parameter increases. When $\lambda = -8$, the loss in each metric is less than 5\%, which is acceptable, and the sparsity can reach 85.3\%. When $\lambda \le -6$, the accuracy loss is unacceptable.
This result demonstrates that the threshold of this method has a certain degree of robustness.

\section{Conclusion}

For sparse attention, we first reformulate the sparsity attention problem as a dual-goal optimization problem and reveal the limitation of the existing methods: the most commonly used Top-K, Top-P cannot fulfill both goals simultaneously. Similarly, we prove that filtering tokens with the maximum weighted by a fixed value is sub-optimal as well. Starting with the optimization problem, we verify that the optimal solution should involve the cumulative energy of the selected tokens and thus propose our \our{} algorithm.
The effectiveness of our method is proven with the results on the Wan 2.2 vision generation model: \our{} reaches 82.14\% sparsity with acceptable accuracy, which is 20.72\% than the BLASST baseline. Moreover, it excels Top-K and Top-P in visual quality under the same sparsity. 
It is also worth noting that this method is closely integrated with the FA algorithm and can be further combined with other methods such as pre-compute masks before FA, feature reuse, and quantization for further acceleration. This method can also be used as a plug-in for existing models and is compatible with post-training to achieve higher accuracy.

\bibliographystyle{plainnat}
\bibliography{biblatex}

\clearpage
\appendix

\section{Appendix / supplemental material}

\subsection{Accuracy vs Recall}
\label{Appd: Accuracy vs Recall}

\begin{figure}[h!]
  \centering

  \includegraphics[width=1\textwidth]{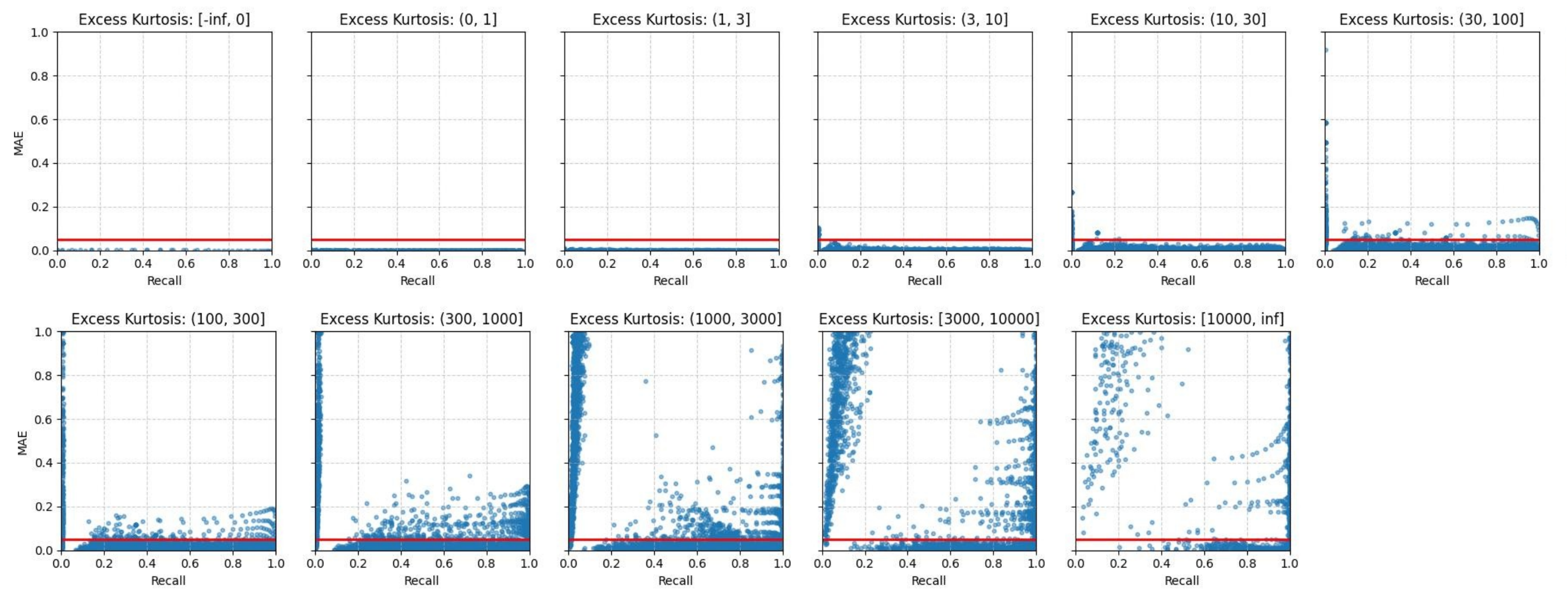}
  \caption{Mean Absolute Error vs Recall rate under different distribution of the $\vP$ matrix.}
  \label{figure:mae-recall}
\end{figure}

In this section, we verify our statements in Sec. \ref{Section:Analysis} with experimental results on Wan 2.2-5B. We sample random rows from the P matrix and apply Top-K with diffent K values to ablate the recall rate.
Excess Kurtosis is used to measure the 'tailedness' of a data distribution. The higher the Excess Kurtosis, the 'peakier' the distribution is.

Fig. \ref{figure:mae-recall} shows how the Mean Absolute Error (MAE) changes with Recall rate under different distributions of the $\vP$ matrix. The results are consistent with our analysis: For flat distribution, a lower recall rate, which is equivalant to a higher $\epsilon$, does not degrade the accuracy significantly. Conversely, as the distribution gets more highly peaked, maintaining a lower $\epsilon$ becomes critical. 

\begin{table*}[h]
  \centering
  \caption{Statistical Analysis of MAE and Recall by Kurtosis Bins}
  \begin{tabular}{c|cc}
    \hline
    Kurtosis Range & MAE at $recall \approx 0.8$ ($\mu \pm \sigma$) & Recall at $MAE \approx 0.05$ ($\mu$ (var)) \\
    \hline
    $[-\infty, 0]$       & $0.0006 \pm 0.0000$         & N/A                       \\
    $(0, 1]$               & $0.0000 \pm 0.0000$         & N/A                       \\
    $(1, 3] $              & $0.0001 \pm 0.0002$         & N/A                       \\
    $(3, 10] $             & $0.0004 \pm 0.0010$         & 0.0007 (0.0000)           \\
    $(10, 30]   $          & $0.0007 \pm 0.0023$         & 0.0272 (0.0038)           \\
    $(30, 100]  $          & $0.0035 \pm 0.0071$         & 0.2087 (0.0814)           \\
    $(100, 300]$           & $0.0066 \pm 0.0153$         & 0.3482 (0.0945)           \\
    $(300, 1000] $         & $0.0060 \pm 0.0112$         & 0.5493 (0.0968)           \\
    $(1000, 3000]$         & $0.0172 \pm 0.0727 $        & 0.6910 (0.0744)           \\
    $[3000, 10000]$        & $0.0325 \pm 0.1745$         & 0.8965 (0.0517)           \\
    $[10000, \infty]$      & $1.9932 \pm 2.4200$       & 0.9386 (0.0201)           \\
    \hline
  \end{tabular}
  \label{tab:sparsity-error var mean}
\end{table*}

Table \ref{tab:sparsity-error var mean} collects the mean and variance of the MAE loss under fixed recall within $0.8\pm0.01$, the recall rate when MAE = $0.05\pm0.005$. As can be seen from the table, when the error threshold is fixed, the recall must be maintained at a relatively high value when the distribution becomes more peaked. However, when the distribution of P becomes flatter, a lower recall can be used without causing significant loss in accuracy.

\subsection{Qualitative Comparison}
Fig. \ref{figure:visualize - frames} shows screenshots of different generation results and the overall sparsity of the video. We controlled the sparsity to be around 80\%, and in terms of image quality, visual consistency, temporal consistency, and realism, \our  demonstrates the best performance.
\begin{figure}[h!]
  \centering
  \includegraphics[width=1.\textwidth]{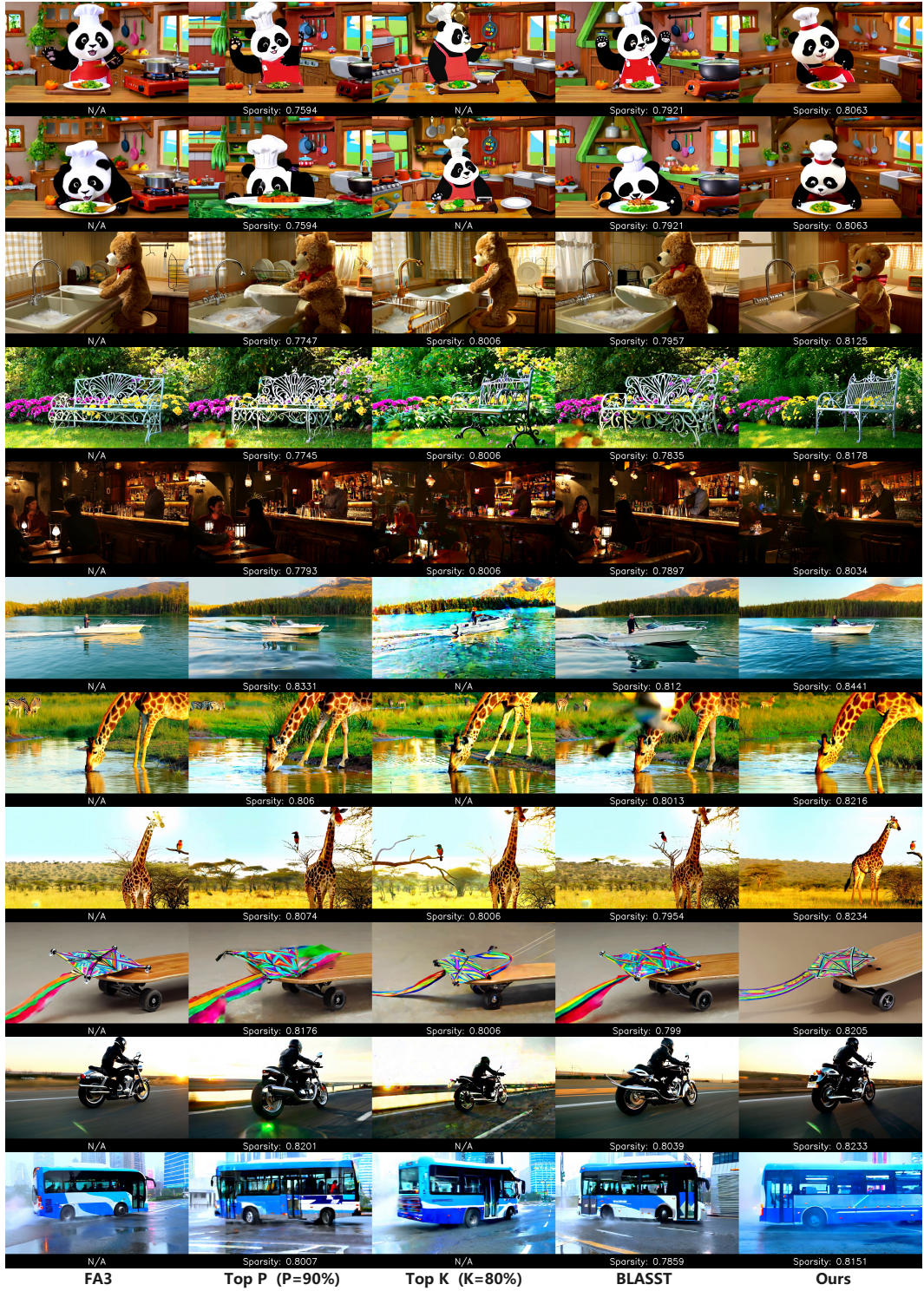}
  \caption{Qaulitative comparison of video frames generated by Wan 2.2 5B with sparsity strategy of Top-P, Top-K, BLASST and \our. Note that the sparsity of some Top-K values is displayed as N/A, and the actual sparsity is 80\%.}
  \label{figure:visualize - frames}
\end{figure}

%%%%%%%%%%%%%%%%%%%%%%%%%%%%%%%%%%%%%%%%%%%%%%%%%%%%%%%%%%%%

\end{document}